\documentclass[10pt,twocolumn]{ICCAS}
 
\usepackage{diagbox}
\usepackage{graphicx}
\usepackage{amsmath}
\usepackage{amsfonts}
\usepackage{caption}
\usepackage{subcaption}
\usepackage{booktabs}
\begin{document}
\pagenumbering{roman}
\title{Bio-Inspired Topological Autonomous Navigation with Active Inference in Robotics}

\author{Daria de Tinguy${}^{1*}$, Tim Verbelen${}^{2}$, Emilio Gamba${}^{3}$, Bart Dhoedt${}^{1}$ }

\affils{ ${}^{1}$Ghent University, Ghent, Belgium \\
(daria.detinguy at ugent.be \\
${}^{2}$Verses, Los Angeles, California, USA, \\
${}^{3}$Flanders Make, Leuven, Belgium, \\
 {\small${}^{*}$ Corresponding author}}


\abstract{Achieving fully autonomous exploration and navigation remains a critical challenge in robotics, requiring integrated solutions for localisation, mapping, decision-making and motion planning. Existing approaches either rely on strict navigation rules lacking adaptability or on pre-training, which requires large datasets. These AI methods are often computationally intensive or based on static assumptions, limiting their adaptability in dynamic or unknown environments.
This paper introduces a bio-inspired agent based on the Active Inference Framework (AIF), which unifies mapping, localisation, and adaptive decision-making for autonomous navigation, including exploration and goal-reaching. Our model creates and updates a topological map of the environment in real-time, planning goal-directed trajectories to explore or reach objectives without requiring pre-training. Key contributions include a probabilistic reasoning framework for interpretable navigation, robust adaptability to dynamic changes, and a modular ROS2 architecture compatible with existing navigation systems. Our method was tested in simulated and real-world environments. The agent successfully explores large-scale simulated environments and adapts to dynamic obstacles and drift, proving to be comparable to other exploration strategies such as Gbplanner, FAEL and Frontiers. This approach offers a scalable and transparent approach for navigating complex, unstructured environments.
}


\maketitle


\section{Introduction}





The transition toward fully automated factories requires robots capable of autonomously exploring and navigating their environments, a key challenge in robotics.

To navigate effectively, an agent must gather and interpret sensory data (e.g., LiDAR, cameras) to perceive its surroundings, adapt to environmental changes, and optimise its motion strategy to minimise computational power and maximise coverage efficiency. Addressing these challenges is essential for achieving robust, scalable, and adaptive robotic navigation in large, dynamic environments.

There are many approaches to tackle the challenges of autonomous navigation. Hard-coded or heuristic-based methods~\cite{frontiers,path_planning_survey} are easily interpretable and computationally cheap but often struggle with dynamic or unstructured environments due to their rigidity. Classical SLAM-based approaches~\cite{orbslam3,gaussiansplattingslam} provide high accuracy in static settings, yet are sensitive to drift and may scale poorly in larger environments. Learning-based methods~\cite{NNslam,BYOL,RECON} perform well in familiar and structured scenarios, but require extensive pre-training and tend to generalise poorly to novel or unexpected conditions. Bioinspired approaches~\cite{ratslam,GSLAM,ours_model} aim to improve adaptability in uncertain settings by modelling navigation through cognitive or topological maps. However, these methods often suffer from computational complexity and have limited deployment in real-world applications.

Despite their strengths, these approaches share limitations: they either depend heavily on prior knowledge, require substantial training, or struggle to adapt in real time to unexpected changes. This motivates the need for frameworks that are both adaptive and data-efficient, capable of operating without pre-training, and resilient in partially observable, dynamic environments.

The Active Inference Framework (AIF), rooted in neuroscience, offers a promising alternative by framing navigation as a predictive inference process. Our model, inspired by AIF principles, operates in a zero-shot, online fashion, continuously learning from incoming sensory data without requiring prior training. This allows it to localise, map, and plan efficiently in environments that are new, visually ambiguous, and dynamic.

Our contributions are as follows.
\begin{itemize}
    \item Novel Perspective in Robotics: We present an integrated model that combines mapping, localisation, and decision-making using Active Inference, enabling robots to navigate without pre-training.
    \item Dynamic Adaptability: The model continuously updates its internal map and beliefs. The agent adapts to changes without requiring preprogrammed rules.
    \item Goal-Oriented Learning: It contains a goal-directed learning mechanism that allows the robot to prioritise exploration or goal reaching, given internal settings.
    \item Modulable Architecture: The system is developed as a set of modules in ROS2, allowing seamless integration with existing robotic platforms and enabling adaptability to various sensor configurations.
\end{itemize}

\vspace{-2mm}
\section{Related works}

Autonomous navigation has been addressed through a range of methods, each targeting aspects like mapping, localisation, and path or motion planning. Traditional methods often rely on hard-coded heuristics or well-established algorithms such as the Dynamic Window Approach or frontier-based exploration~\cite{frontiers,DWA}, offering computational efficiency and interpretability but lacking adaptability in dynamic or unstructured settings~\cite{path_planning_survey}.

Metric SLAM-based systems~\cite{orbslam3,gaussiansplattingslam,FAST-LIO2} remain a cornerstone in robotics; they provide precise localisation and mapping, particularly in static environments. However, they are prone to drift and may scale poorly with environmental complexity. To overcome these limitations, bio-inspired topological methods like RatSLAM~\cite{ratslam} and G-SLAM~\cite{GSLAM} use graph-based representations, which are memory-efficient and robust to perceptual aliasing. Despite this, they often struggle in dynamic environments due to their reactive nature.
Learning-based techniques such as Neural-SLAM~\cite{NNslam}, RECON~\cite{RECON}, and BYOL-Explore~\cite{BYOL} perform well in structured, familiar settings but require extensive pre-training and large datasets, limiting generalisability. These systems often act as black boxes, complicating interpretability and online adaptation.
Meanwhile, model-based control strategies using Model Predictive Path Integral (MPPI)~\cite{AGVs,MPPI} enable accurate planning but are computationally intensive and reliant on precise models. Hybrid methods combining planning with learned priors~\cite{ETPNav} show promise but face challenges in real-time, general-purpose navigation.
To address these limitations, we propose a model based on the Active Inference Framework (AIF)~\cite{AIF_book}, which unifies mapping, localisation, and planning through predictive belief updating. Active inference treats navigation as a continual process of prediction and belief updating, minimising surprise about the environment. This leads to agents that are both reactive and anticipatory, capable of adapting in real time to ambiguous conditions. Although AIF has been applied conceptually to navigation~\cite{nav_aif,curiosity_exploitative}, there are few implementations in robotics~\cite{AIF_robot_nav,GSLAM}. These typically depend on past observations and/or pre-training, making them sensitive to environmental changes and less suited for exploring unknown areas.

Our model extends this by constructing a topological map that combines past experiences with forward predictions of unexplored regions~\cite{ours_model}. It can infer possible navigable spaces (e.g., behind doorways) and plan accordingly without prior training. With its zero-shot, online learning design, the model continuously updates its internal representation based on real-time sensory input. This compositional and interpretable structure allows robust navigation in dynamic, partially observable environments while supporting modularity and real world deployment. 
\vspace{-5mm}
\section{Method}
Our method enables an autonomous agent to build and navigate a cognitive map of its environment using a probabilistic generative model. Inspired by Active Inference, the system continuously updates its internal representation based on partial sensory observations, inferred beliefs, and expected outcomes of future actions. This section describes the core components of our model, including environment mapping, map expansion, decision-making, and obstacle handling. We also detail the system's architecture and implementation within the ROS2 framework, highlighting how its modular structure supports generalisation across different robotic platforms and sensor configurations.
\subsection{Mapping the Environment}
A cognitive map refers to an internal representation of spatial knowledge, enabling agents to navigate and interpret their surroundings~\cite{cscg_pres,cscg_structuring_knowledge,humans-mapping,humans-cognitive-map}. In robotics, this concept aligns with topological graphs. Our model builds such a representation as a graph, where each node corresponds to a possible state defined by the agent’s believed pose and sensory observations. State certainty depends on how well current observations match the agent’s belief. This topological structure allows for memory-efficient scalability to large environments. To manage the computational cost of long-term action prediction, we combine Active Inference with a Monte Carlo Tree Search algorithm~\cite{mcts}.

\begin{figure}
    \centering
    \includegraphics[width=0.5\linewidth]{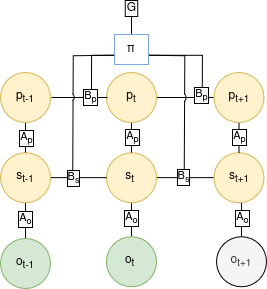}
    \caption{Factor graph of the POMDP in our generative model, showing transitions from the past to the present (up to time-step $t$) and extending into the future (time-step $t+1$). Past observations are marked in green, indicating they are known. In the future steps, actions follow a policy $\pi$ influencing the new states and position in yellow and new predictions in grey. The position at time $t$, $p_t$, is determined by the policy and the prior position $p_{t-1}$, while the current state $s_t$ is inferred from the observation $o_t$, the position $p_t$, and the previous state $s_{t-1}$. Transitions between states are ruled by the $B$ matrices, which define how prior conditions contribute to the current one, considering taken actions. $A$ matrices represent the probabilities of an observation corresponding to a state.}
    \label{fig:pomdp}
    \vspace{-2mm}
\end{figure}

To improve the internal map of the agent, representing how the robot can navigate the environment, our model adjusts its parameters to minimise Free Energy, a concept in Active Inference (AIF)~\cite{AIF_book}. AIF posits that action and perception aim to minimise an agent's Free Energy, acting as an upper limit to surprise. Generative models are central to active inference, as they encapsulate causal relationships among observable outcomes, agent actions, and hidden environmental states in a latent space. These environmental states remain 'hidden' as they are shielded from the agent's internal states by a Markov blanket~\cite{World_Model}. Leveraging partial observations, the agent constructs its own beliefs regarding hidden states, enabling action selection and subsequent observation to refine its beliefs relying on the Partially Observable Markov Decision Model (POMDP)~\cite{AIF_learning}. Our POMDP is presented Figure~\ref {fig:pomdp}.

The active inference framework also defines how to update model parameters in response to new evidence in a changing environment. The generative model of our model is defined in Equation~\eqref{eq:pomdp}, where the current state $s_t$ and position $p_t$ are inferred based on the previous state $s_{t-1}$, position $p_{t-1}$ and action $a_{t-1}$ leading to the current observation $o_t$. The joint probability distribution $P$ is defined over time sequences of states, observations, and actions with Tildes~(~$\tilde{}$~) denoting sequences over time. 
\begin{equation} 
\begin{aligned}
P(\tilde{o}, \tilde{s}, \tilde{p} ,\tilde{a})  = &  P(o_0| s_{0})P(s_0)P(p_0) \\
&\prod_{t=1}^\tau 
P(o_t| s_{t})P(s_t, p_t|s_{t-1},p_{t-1},a_{t-1})  
\label{eq:pomdp}
\end{aligned}
\end{equation}

Our model works with the following essential distributions:
\begin{itemize}
    \item State transitions ($B_s$): Likelihood of the agent moving between states.
    \item Position transitions ($B_p$): Likelihood to move between positions.
    \item Observation likelihoods ($A_o$): How likely certain observations are at each state.
    \item Position likelihoods ($A_p$): The likelihood of being in a particular position at each state.
\end{itemize}

The states and positions are inferred at each time-step $t$, while the agent perceives the observations. Those elements are defined as Markov matrices in the agent, except for $B_p$, which is a tensor containing poses (tuples containing believed x and y coordinates) incrementally increasing as new ones are added. $B_p$ is a key element to map extension, as we will explain later.     

\begin{figure}[htb!]
\centering    
\begin{subfigure}[t]{0.49\columnwidth}
    \centering
    \includegraphics[width=\textwidth]{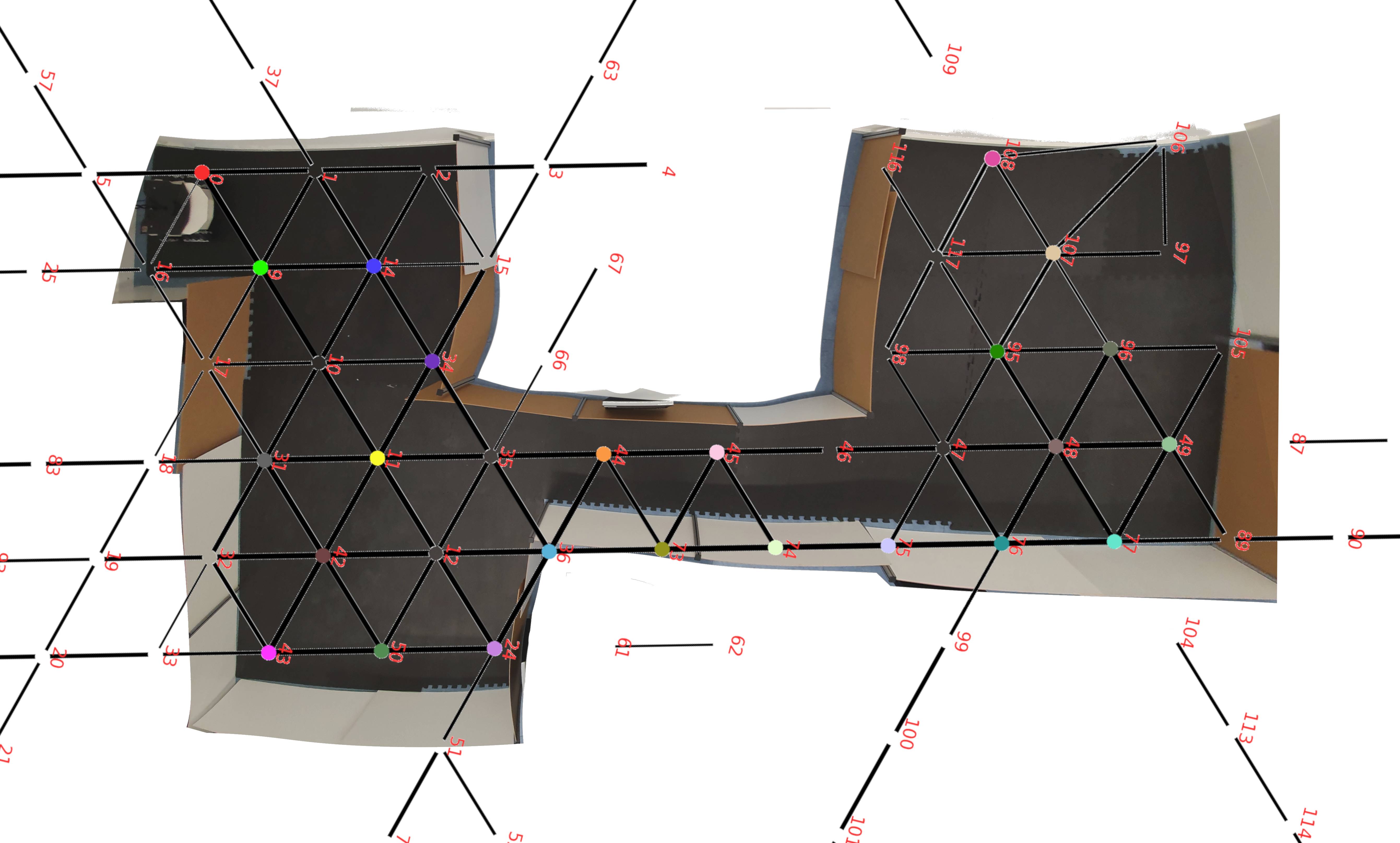}
    \caption{5m$^2$ of real environment}
    \label{fig:map_maze_rw}
\end{subfigure}
\hfill
\begin{subfigure}[t]{0.49\columnwidth}
    \centering
    \includegraphics[width=\textwidth]{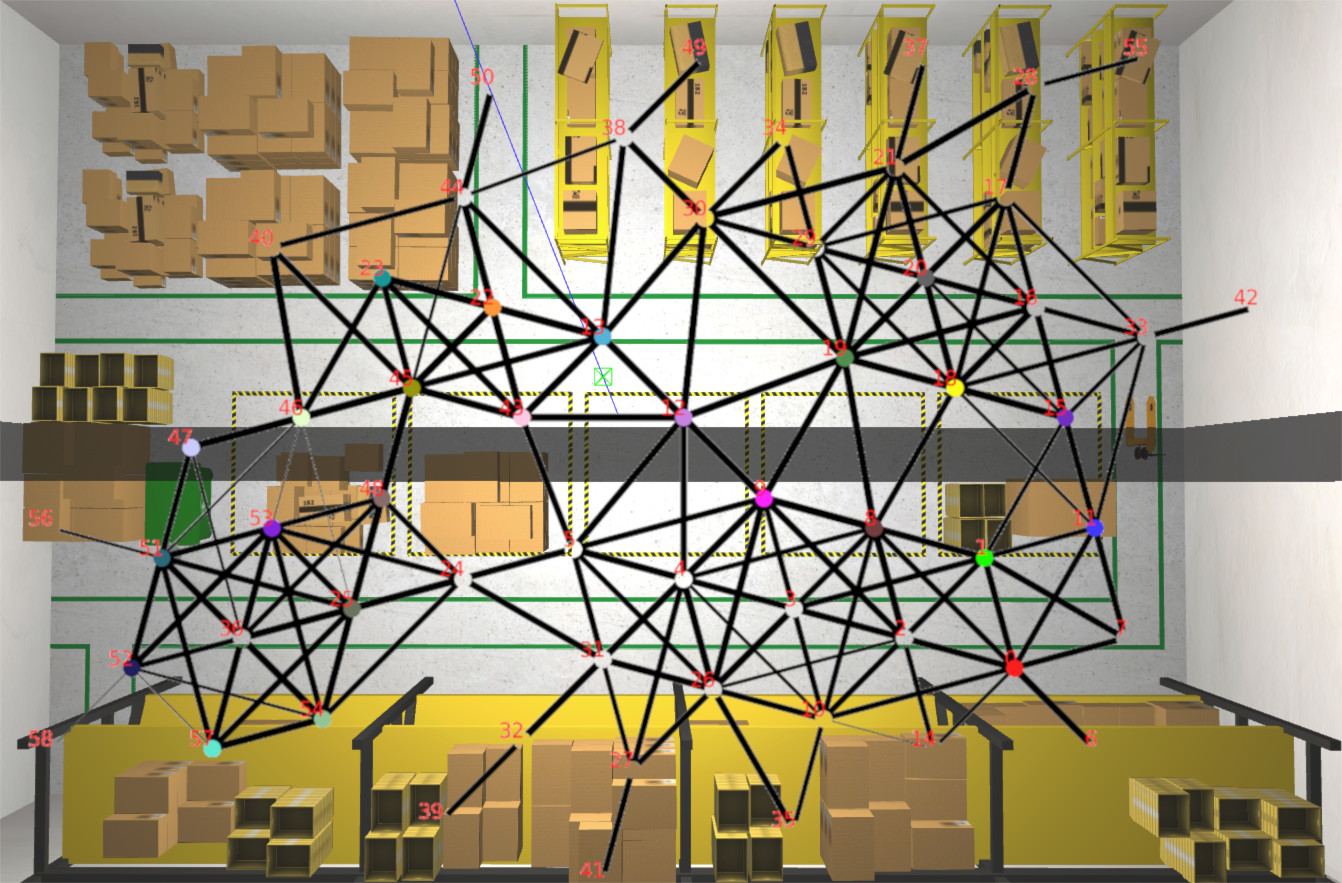}
    \caption{280m$^2$ simulated warehouse}
    \label{fig:map_big_ware}
\end{subfigure}
\caption{Final map of exploration in a) a real-world environment, b) Amazon simulated warehouse. Coloured points signify visited locations, where the same colour attributions mean the same observation. The thickness of the lines depicts the agent's believed probability of transitioning between two states given an action.}
\label{img:maps}
\vspace{-2mm}
\end{figure}

Our model maps according to a given influence radius (the minimum acceptable distance between two locations), which the user can freely define depending on their environment. Figure ~\ref{img:maps} presents the agent's generated map in a small real maze of 5m$^2$and over a fully mapped simulated warehouse of 280m$^2$~\cite{warehouse}.   

\subsection{Expanding the Map}
The agent continuously updates its topological map and visual memory by integrating new sensory inputs and motion predictions. At each new motion, it expands the set of locations in $B_p$ by identifying accessible areas (currently based on the LiDAR range, although this could be replaced by visual cues such as door recognition~\cite{ghost_node_explo}). This process dynamically expands relevant matrices according to the computed Expected Free Energy (EFE) over policies.

To recognise locations, in this specific architecture, the agent builds a 360° panorama from stitched images and compares it to stored memories using a Structural Similarity (SSIM) threshold. 
If no match is found, it may either indicate a new place or an environmental change. The agent uses its believed state $s$, inferred from estimated position $p$ and current observation $o$, to determine how to update the model; the confidence in $s$ is verified. Suppose confidence in $s$ exceeds a set threshold. In that case, it assumes correct localisation and the likelihood matrix $A_o$ grows to receive this observation and associate it with the present state, which makes it resistant to visual changes such as lighting and perceptual aliasing (two similar images at different locations).
If localisation confidence is lower than the set threshold, the agent prioritises re-localising by searching for familiar consecutive views before recording changes and updating the model.

For planning, the agent uses its current pose and obstacle data (e.g., LiDAR) to predict the outcome of actions and updates its internal map accordingly. New unexplored regions are added as hypothetical nodes (position $p$, observation $o$ unknown until visited). EFE guides action selection via a softmax over the model likelihood $A_p$, based on:

\begin{itemize}
    \item Expected information gain: how well pose $p$ explains the environment, considering collision likelihood $c$ from $o$.
    \item  Expected value: probability of getting a desired $o$ and encountering a collision $c$.
\end{itemize}  

These factors jointly drive decision-making under Active Inference. Locations with uncertain observations naturally attract exploration, as they maximise information gain.
\subsection{Decision Making} 
The agent navigates by selecting among discrete orientation ranges spanning the full 360$\deg$ yaw (e.g., [0–30], [30–60], etc.). The definition of these angular segments, as well as the inclusion of a "STAY" action, is user-configurable.
The model uses EFE to guide its navigation decisions. EFE minimises expected surprise by favouring policies increasing the likelihood of encountering preferred states. Those preferences can be given as a specific observation (utility term) or the curiosity to comprehend the environment's structure~\cite{curiosity_exploitative} (information gain term). This approach facilitates active learning by rapidly reducing uncertainty about model parameters and enhancing knowledge acquisition about unknown contingencies. 

At the heart of decision-making and adaptive behaviour lies a delicate balance between exploitation and exploration~\cite{curiosity_exploitative}. Exploitation involves selecting the most valuable option based on existing beliefs about the world, while exploration entails choosing options to learn and understand the environment~\cite{world_model_and_inference}.

When the agent predicts new states and updates its map, depending on our agent's objective and the weight between exploitation (reaching a desired objective) and exploration (learn about the environment), the agent will prioritise locations that provide the most interesting information.
For a rigorous mathematical treatment of the AIF model used here, we refer the reader to~\cite{ours_model}. The present paper focusing  more on the application of an AIF model to robotics.
\subsection{Handling dynamic obstacles}
The agent also deals with obstacles during navigation. When an obstacle is detected by the lidar (like a wall or box) or experimented (the agent can not move toward the objective) the model updates its transition probabilities between two states, decreasing the likelihood of moving into that blocked space using a Dirichlet pseudo-count update defined in equation~\eqref{eq:B_up}, with a positive or negative learning rate $\lambda$ depending on the situation presented on table~\ref{tab:tran_lr}. 

\begin{equation}
    B_\pi = B_\pi + Q(s_t|s_{t-1}, \pi) Q(s_{t-1}) * B_\pi  * \lambda
    \label{eq:B_up}
\end{equation}

\begin{table}[h!]
\centering
\caption{Transition learning rate ($\lambda$) depending on the situation}
\resizebox{\columnwidth}{!}{
\begin{tabular}{lcccc}
\toprule
\textbf{Transitions} & \textbf{Possible} & \textbf{Impossible} & \shortstack{\textbf{Predicted} \\ \textbf{Possible}} & \shortstack{\textbf{Predicted} \\ \textbf{Impossible}} \\
\midrule
\hline
Forward  & 7  & -7 & 5 & -5 \\\hline
Reverse  & 5  & -5 & 3 & -3 \\
\hline
\bottomrule
\end{tabular}}
\label{tab:tran_lr}
\end{table}

\begin{figure}[h!]
\centering    
\begin{subfigure}[t]{0.45\columnwidth}
    \centering    \includegraphics[width=\textwidth]{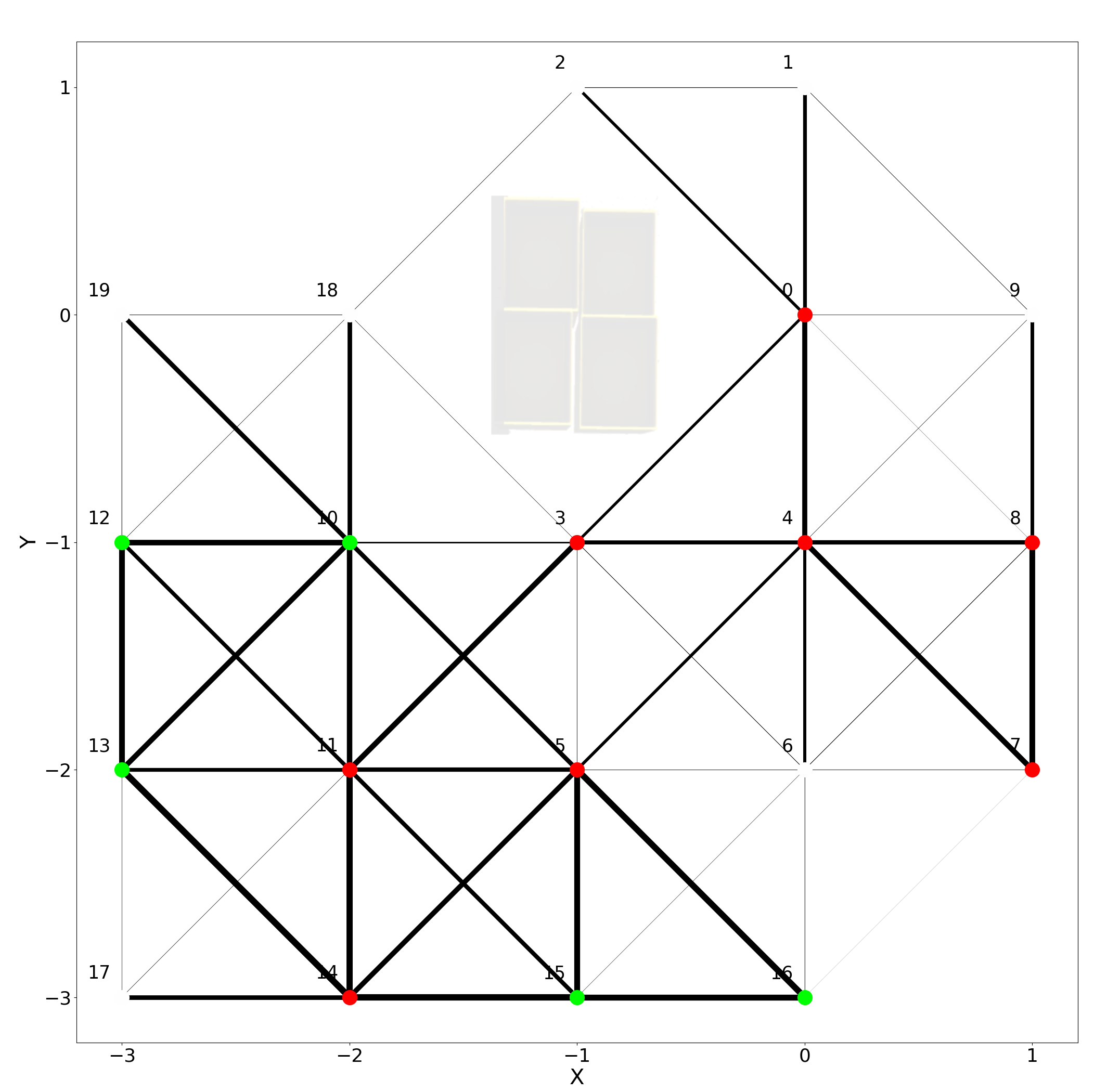}
    \caption{Agent map with an obstacle initially placed at position (-1,0) before moving it after a partial exploration.}
    \label{img:ob_21_steps}
\end{subfigure}\hfill
\begin{subfigure}[t]{0.45\columnwidth}
    \centering
    \includegraphics[width=\textwidth]{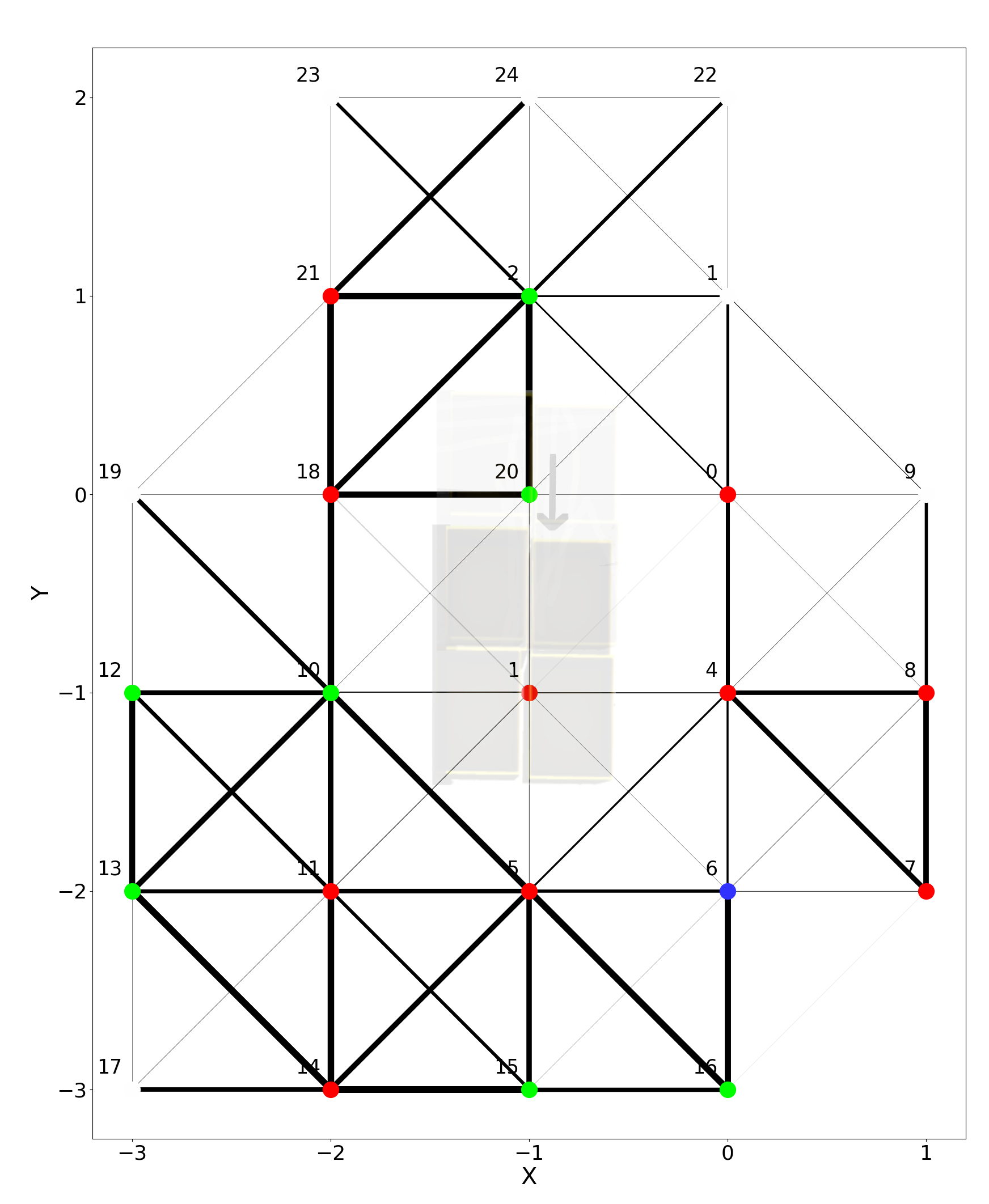}
    \caption{Obstacle moved at position (-1,-1). After 20 more steps, the failure to reach the covered state reduces transition probabilities, represented by thinner lines.}
    \label{img:ob_40_steps}
\end{subfigure}
\caption{Obstacle movement and its impact on the agent's internal map during exploration.}
\vspace{-2.5mm}
\label{img:ob_move}

\end{figure}

Thus, our model can predict obstacles, and if an obstacle was not detected (e.g., the LiDAR failed to pick it up), it can still recover from a failed motion. The learning rate of the transition matrix between two states depends on the situation, whether the model is predicting a blockage or experiencing one. Predicted motions (feasible or not) have a lower impact on the learning rate of transitions compared to transitions that the model has directly experimented through motion. Figure~\ref{img:ob_move} exemplifies this process with an obstacle (e.g. a box) moved between two positions (from position (-1,0) to (-1,-1) over the visited state number 3) while the agent explored a clustered environment of 36m$^2$~\cite{warehouse} with 8 action ranges to chose from. 

As the agent fails to reach state number 3 after the box covers it, the model adjusts its internal map to reflect this new reality. The transition probabilities to that location reduce, and the state ID at this position becomes incorrect (state 1 instead of 3) because the agent cannot correct its belief by moving to that location. A new state (state 20) is created at the former position of the obstacle, and new transitions are established. This change does not affect any of the other existing states.
\subsection{Agent Architecture}
With our method established, we now turn to its implementation and evaluation. To assess our model’s performance in realistic scenarios, we integrate it into the Robot Operating System (ROS2)~\cite{ros2}. The system is imagined so that each module is independent and can be improved or replaced independently. As such, this work focuses on the model map generation and planning capacity rather than the visual module efficiency or motion planning flexibility.  
Our model is robot and sensor agnostic and was adapted to several robots (turtlebot, turtlebot3~\cite{turtlebots} and rosbotxl~\cite{rosbotxl}) with forward or 360$\deg$ lidars of various ranges, with single, several cameras or a 360$\deg$ camera. 

\begin{figure}[bht!]
    \centering
    \includegraphics[width=0.75\columnwidth]{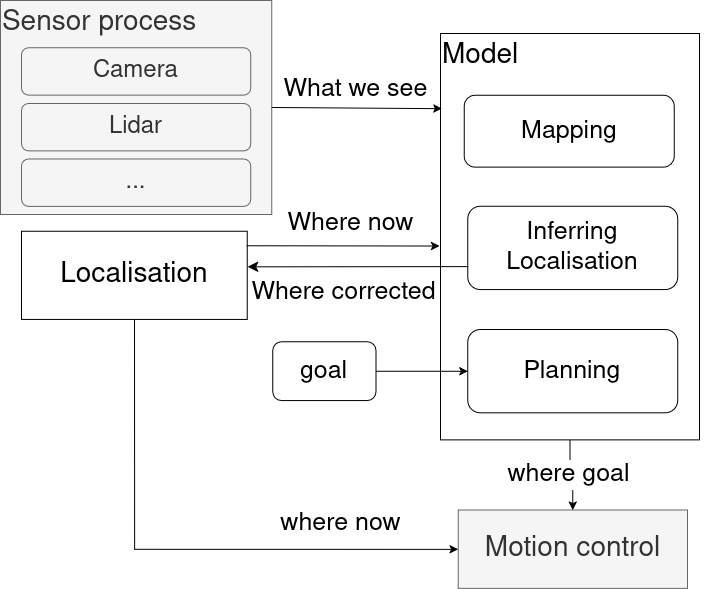}
    \caption{Overview of the system architecture. In grey, we have modules that any ROS-compatible solution can replace. Modules interact through belief propagation, Inferring and planning (localisation, mapping and action selection) rely on the AIF framework. The perceptual and motion planning still use traditional approaches. Believed odometry takes precedence over sensor odometry. Preferences (goal) are expected from the user if we want to reach a target observation.}
    \label{fig:archi}
    \vspace{-5mm}
\end{figure}

Our system comprises four modules (Figure~\ref{fig:archi}, grey modules are meant to be replaced by any ROS-compatible modules): the model, odometry, sensor processing, and motion control. The model's planning relies on believed odometry (estimated internally by the model) rather than sensor-based data, avoiding issues like drift but resulting in approximate positioning. The agent navigates not to precise coordinates, but to locations expected to produce desired observations. Motion control uses either Nav2~\cite{nav2} or a potential field~\cite{potential_field} to move between locations defined by the model. The observation module captures panoramic views at specific locations and matches them to past experiences before sending a belief-based observation to the model. While currently simple, this sensor module is designed for future upgrades, particularly in observation and place recognition.


\vspace{-2.5mm}
\section{Results}
Having outlined the architecture and design of our model, we now move to evaluate its performance in various environments with diverse features and dimensions. 
\subsection{Exploring various environments}
We evaluate our model in four simulated environments: a mini (36m$^2$), small (80m$^2$) and large (280m$^2$) warehouse, inspired by the Amazon Gazebo environment~\cite{warehouse}, featuring aisles, boxes, and industrial obstacles like forklifts; and a 156m$^2$ home environment~\cite{house} without doors, including a kitchen-living area, sports and play rooms, and a bedroom. Both settings include challenging objects for LiDAR-based detection, such as curved chairs or forklifts.

In each scenario, the agent begins exploration from random initial positions. Since our model does not construct a metric map, we rely on LiDAR range readings to associate observations with the agent’s internal representation of new state locations.

\begin{figure}[ht!]
    \centering
\includegraphics[width=0.82\columnwidth]{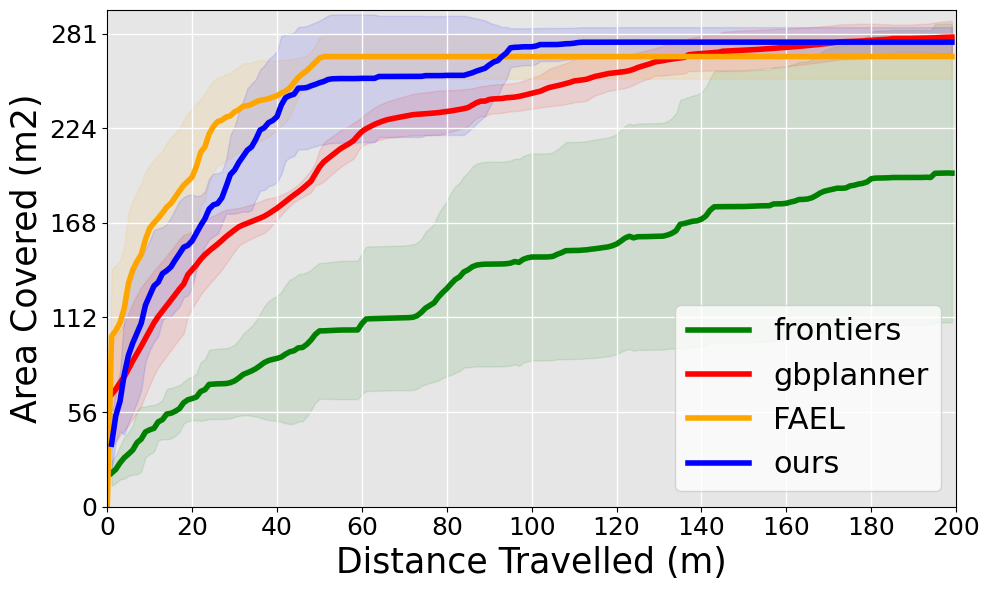}
    \caption{Exploration efficiency in distance over the whole area of the warehouse by our model, Frontiers, FAEL and Gbplanner.}
    \label{fig:coverage_big_ware}
\end{figure}

The agent explored the 280m$^2$ warehouse from five starting points using four different approaches: (1) our proposed method combining mapping, localisation, and planning with Nav2 for motion control; (2) the Frontier-based exploration algorithm~\cite{frontiers}, implemented with Nav2 SLAM~\cite{nav2_slam}; (3) FAEL~\cite{FAEL} based on Frontiers, creating a 3D map using UFOMap~\cite{ufomap}, and a topological graph to navigate; and (4) Gbplanner~\cite{gbplanner}, an enhanced version of the 2021 DARPA SubT Challenge winner~\cite{Darpa_winners}, which leverages a Voxblox-based~\cite{voxblox} 3D map for topological planning.

Sensor setups varied across methods: both our model and Frontiers used a 12 m-range 2D 360° LiDAR and a single camera, whereas Gbplanner operated with three cameras and two 12 m-range 3D LiDARs. All experiments were performed using a Turtlebot3 Waffle robot.

Figure~\ref{fig:coverage_big_ware} shows the exploration efficiency of each model to fully scout the environment (averaged over 5 runs). Frontiers, lacking optimised navigation, required multiple passes over the same areas. Especially when some small, unreachable areas have unexplored zones without a clear frontier. In contrast, our model and Gbplanner follow a more efficient trajectory (see an example of our trajectory in Figure~\ref{fig:drift}). This displays that our bio-inspired approach is on par with optimised exploration strategies. 




\subsubsection{Obstacle robustness}
Gbplanner and our model both support dynamic replanning, though our approach reacts immediately to changes, even those occurring directly ahead of the agent. Gbplanner only replans after a few steps toward its objective, so it might not be able to react to an obstacle appearing right in front of it. Frontiers theoretically adapts through its use of Nav2 SLAM, but it often struggles in practice due to reliance on feature-based localisation and a static map planner. It results in our model being more apt to recover from encountering a surprising or undetectable obstacle (e.g., a curved chair base or a forklift). In such cases, other models failed to account for the stationary motion and require an external intervention. Table~\ref{tab:human_int} summarises the number of human interventions required per environment over all runs. In all cases, the interventions for our agent were due to physical limitations (e.g., the two robot wheels lifted off the ground after hitting an obstacle), while other models also required interventions due to an inability to consider the obstacle and replan. FAEL also got stuck in open areas and required a little push to correctly consider Lidar data. 

\begin{table}[tbh!]
\centering
\caption{Number of times the robot got stuck and required intervention per environment over all runs.}
\begin{tabular}{l|cccc}
\begin{tabular}[c]{@{}l@{}}External \\ intervention\end{tabular} & ours    & Gbplanner  & Frontiers & FAEL \\ \hline
Mini ware & 0 & 2 & 3 & 3         \\ \hline
Small ware   & 1 & 4 & 2 & 2       \\ \hline
Big ware  & 0 & 2 & 0 & 5        \\ \hline
Home   & 2 & 2 & 5 & 5       
\end{tabular}
\label{tab:human_int}
\vspace{-1mm}
\end{table}

Overall, both our model and Gbplanner demonstrated efficient exploration strategies, covering a 280m$^2$ environment in approximately 100m travelled. Our model exhibited great robustness by maintaining exploration efficiency across different starting points and recovering from undetectable obstacles.

\subsubsection{Drift robustness} 
\begin{figure}[ht!]
    \centering
    \includegraphics[width=0.7\linewidth]{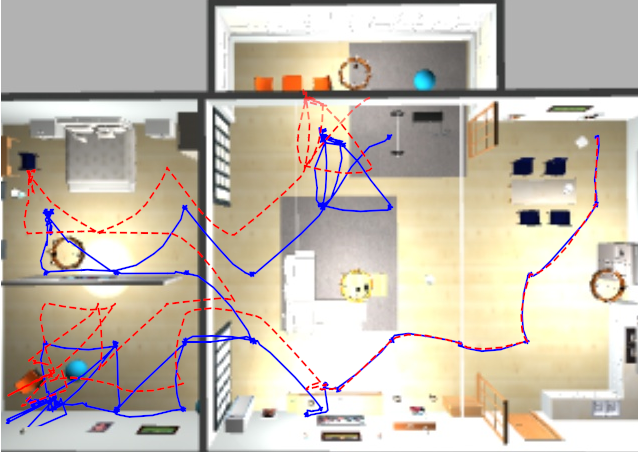}
    \caption{Drift between the real odometry (dashed red) and the agent-believed odometry (continuous blue) during a home exploration. The drift is due to a furniture collision and location approximation.}
    \label{fig:drift}
    \vspace{-2mm}
\end{figure}

In one of the home runs, depicted in figure~\ref{fig:drift}, the agent drifted between the real odometry and the believed odometry of the agent due to getting stuck on furniture and approximating its localisation. With our model, the exploration and goal-reaching ability is not impacted by this shift, showing the robustness of the model to drift and exploration efficiency. Frontiers had a similar experience with an object that the lidar could not detect, the odometry shifted, and nav2 SLAM was not able to recover from this. The used Frontiers algorithm~\cite{frontiers} and Gbplanner~\cite{gbplanner} do not propose an alternative objective to reach if navigation fails. 
\subsection{Reaching given objective}
The model can be given goals either as positions or RGB observations. It first determines whether the goal is familiar, then either explores or navigates directly to it. In these trials, RGB observations were used, and the agent had already explored the environment, allowing comparison between the agent's path and the ideal trajectory.

\begin{figure}[ht!]
    \centering
    \includegraphics[width=0.77\linewidth]{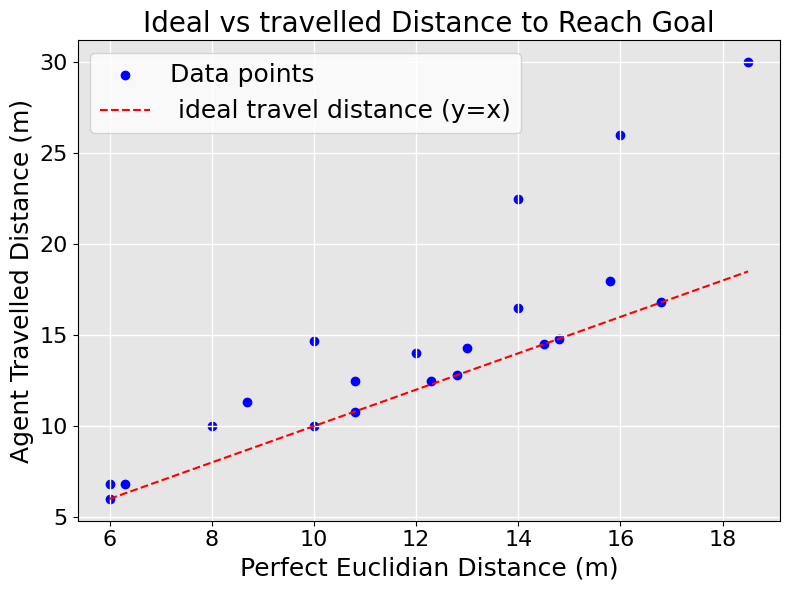}
    \caption{Euclidean distance and navigated distance from the starting position to the goal through all environments.}
    \label{fig:steps_comparison}
    \vspace{-5mm}
\end{figure}

At each step, the agent estimates its position and is allowed to imagine trajectories up to around 14 m to select the optimal action. To prioritise reaching the goal, the exploration term was set to 0, and the preference for the target observation was weighted at 5. This encourages efficient, goal-directed behaviour without distraction from unexplored areas.
Results show that the agent reliably follows near-optimal paths, with minimal deviation up to 14 meters. Beyond its imagination horizon (i.e. the range within which it predicts the consequences of its actions), it may briefly explore before recalculating a direct route. This is illustrated in Figure~\ref{fig:steps_comparison}.

This limitation can be mitigated by applying strategies such as~\cite{goal_grid_cell} or \cite{inductive_AIF}, where a weaker preference or a state preference is propagated from the goal to connected nodes, guiding the agent when the goal is over the prediction range. We could also develop sub-goals as~\cite{overcome_explo} propose in their deep learning navigation model. Additionally, the RGB observation could be associated with a specific position, allowing the agent to assign weight to both the preferred position and the observation.
The robots equipped with our growing cognitive map and an active inference navigation scheme can explore a dynamic environment with no human intervention, successfully adapting to changes and efficiently mapping their surroundings. These results indicate that our approach has the potential to be applied to industrial scenarios. By having adaptive agents following principles inspired by animal navigation, robots could learn and explore in environments where rigid, hard-coded strategies would fail. We envision that such robots could autonomously operate in large real-world environments, such as warehouses, where conditions are often variable without pre-training or strictly defined navigation rules. 
\vspace{-6mm}
\section{Discussion}
Our model uses Active Inference (AIF) to build a cognitive map by integrating visual observations and estimated body position into a generative topological graph~\cite{build_cogn_map}. Unlike existing AIF systems that rely solely on past observations~\cite{AIF_modeling_struct,bayesian_model_reduc}, our approach predicts possible future states, enabling proactive and adaptive exploration~\cite{ours_flex_Cmap}. This results in a robust, self-organising internal map that guides decision-making without requiring prior training, offering zero-shot, online learning capabilities.
The framework is modular, based on ROS2~\cite{ros2} and fully interpretable, with decision-making grounded in the minimisation of expected free energy~\cite{nav_aif}. 
This explainability allows users to trace how navigation decisions are made and how predictions shape exploration or goal-reaching.

Our model proves to be as reliable in exploration tasks as state-of-the-art (GBplanner~\cite{gbplanner}, FAEL~\cite{FAEL}) topological planning and more efficient than traditional methods such as Frontiers~\cite{frontiers}, even in dynamic environments. Future enhancements like semantic perception~\cite{object_nav_semantic} or hierarchical planning~\cite{ours_hierarchy} could improve generalisation and efficiency, and are meant to be integrated. More tests should be conducted in a larger real-world warehouse for exploration and goal-reaching to fully confirm those results. Moreover, we still have to formally demonstrate that our method is computationally affordable in robotics platforms. Despite these limitations, our method demonstrated that a biologically inspired, modular, and adaptable navigation system is possible using AIF, which is promising for real-world robotic applications in uncertain and changing environments.

\section*{ACKNOWLEDGEMENT}
This research received funding from the Flemish Government under the “Onder-zoeksprogramma Artificiële Intelligentie (AI) Vlaanderen” programme.

%

%
\bibliographystyle{ieeetr}
\bibliography{main}

\begin{thebibliography}{10}

\bibitem{frontiers}
A.~Topiwala, P.~Inani, and A.~Kathpal, ``Frontier based exploration for autonomous robot,'' 2018.

\bibitem{path_planning_survey}
H.~S. Hewawasam, M.~Y. Ibrahim, and G.~K. Appuhamillage, ``Past, present and future of path-planning algorithms for mobile robot navigation in dynamic environments,'' {\em IEEE Open Journal of the Industrial Electronics Society}, vol.~3, pp.~353--365, 2022.

\bibitem{orbslam3}
C.~Campos, R.~Elvira, J.~J. Gomez, J.~M.~M. Montiel, and J.~D. Tardos, ``{ORB-SLAM3}: An accurate open-source library for visual, visual-inertial and multi-map {SLAM},'' {\em IEEE Transactions on Robotics}, vol.~37, no.~6, pp.~1874--1890, 2021.

\bibitem{gaussiansplattingslam}
H.~Matsuki, R.~Murai, P.~H.~J. Kelly, and A.~J. Davison, ``Gaussian splatting slam,'' 2024.

\bibitem{NNslam}
D.~S. Chaplot, D.~Gandhi, S.~Gupta, A.~Gupta, and R.~Salakhutdinov, ``Learning to explore using active neural slam,'' in {\em International Conference on Learning Representations (ICLR)}, 2020.

\bibitem{BYOL}
Z.~D. Guo, S.~Thakoor, M.~Pîslar, B.~A. Pires, F.~Altché, C.~Tallec, A.~Saade, D.~Calandriello, J.-B. Grill, Y.~Tang, M.~Valko, R.~Munos, M.~G. Azar, and B.~Piot, ``Byol-explore: Exploration by bootstrapped prediction,'' 2022.

\bibitem{RECON}
D.~Shah, B.~Eysenbach, G.~Kahn, N.~Rhinehart, and S.~Levine, ``Rapid exploration for open-world navigation with latent goal models,'' 2023.

\bibitem{ratslam}
M.~Milford, G.~Wyeth, and D.~Prasser, ``Ratslam: a hippocampal model for simultaneous localization and mapping,'' in {\em IEEE International Conference on Robotics and Automation, 2004. Proceedings. ICRA '04. 2004}, vol.~1, pp.~403--408 Vol.1, 2004.

\bibitem{GSLAM}
A.~Safron, O.~Çatal, and T.~Verbelen, ``Generalized simultaneous localization and mapping (g-slam) as unification framework for natural and artificial intelligences: towards reverse engineering the hippocampal/entorhinal system and principles of high-level cognition,'' {\em Frontiers in Systems Neuroscience}, vol.~Volume 16 - 2022, 2022.

\bibitem{ours_model}
D.~de~Tinguy, T.~Verbelen, and B.~Dhoedt, ``Learning dynamic cognitive map with autonomous navigation,'' {\em Frontiers in Computational Neuroscience}, vol.~18, Dec. 2024.

\bibitem{DWA}
D.~Fox, W.~Burgard, and S.~Thrun, ``The dynamic window approach to collision avoidance,'' {\em IEEE Robotics and Automation Magazine}, vol.~4, no.~1, pp.~23--33, 1997.

\bibitem{FAST-LIO2}
W.~Xu, Y.~Cai, D.~He, J.~Lin, and F.~Zhang, ``{FAST-LIO2:} fast direct lidar-inertial odometry,'' {\em CoRR}, vol.~abs/2107.06829, 2021.

\bibitem{AGVs}
I.~S. Mohamed, K.~Yin, and L.~Liu, ``Autonomous navigation of agvs in unknown cluttered environments: Log-mppi control strategy,'' {\em IEEE Robotics and Automation Letters}, vol.~7, no.~4, pp.~10240--10247, 2022.

\bibitem{MPPI}
I.~S. Mohamed, J.~Xu, G.~S. Sukhatme, and L.~Liu, ``Towards efficient mppi trajectory generation with unscented guidance: U-mppi control strategy,'' 2024.

\bibitem{ETPNav}
D.~An, H.~Wang, W.~Wang, Z.~Wang, Y.~Huang, K.~He, and L.~Wang, ``Etpnav: Evolving topological planning for vision-language navigation in continuous environments,'' 2024.

\bibitem{AIF_book}
T.~Parr, G.~Pezzulo, and K.~Friston, {\em Active Inference: The Free Energy Principle in Mind, Brain, and Behavior}.
\newblock The MIT Press, 03 2022.

\bibitem{nav_aif}
R.~Kaplan and K.~Friston, ``Planning and navigation as active inference,'' {\em bioRxiv}, 12 2017.

\bibitem{curiosity_exploitative}
P.~Schwartenbeck, J.~Passecker, T.~U. Hauser, T.~H. FitzGerald, M.~Kronbichler, and K.~J. Friston, ``Computational mechanisms of curiosity and goal-directed exploration,'' {\em eLife}, vol.~8, p.~e41703, may 2019.

\bibitem{AIF_robot_nav}
O.~Çatal, T.~Verbelen, T.~{Van de Maele}, B.~Dhoedt, and A.~Safron, ``Robot navigation as hierarchical active inference,'' {\em Neural Networks}, vol.~142, pp.~192--204, 2021.

\bibitem{cscg_pres}
D.~George, R.~Rikhye, N.~Gothoskar, J.~S. Guntupalli, A.~Dedieu, and M.~Lázaro-Gredilla, ``Clone-structured graph representations enable flexible learning and vicarious evaluation of cognitive maps,'' {\em Nature Communications}, vol.~12, 04 2021.

\bibitem{cscg_structuring_knowledge}
M.~Peer, I.~K. Brunec, N.~S. Newcombe, and R.~A. Epstein, ``Structuring knowledge with cognitive maps and cognitive graphs,'' {\em Trends in Cognitive Sciences}, vol.~25, no.~1, pp.~37--54, 2021.

\bibitem{humans-mapping}
P.~Foo, W.~Warren, A.~Duchon, and M.~Tarr, ``Do humans integrate routes into a cognitive map? map- versus landmark-based navigation of novel shortcuts.,'' {\em Journal of experimental psychology. Learning, memory, and cognition}, vol.~31, pp.~195--215, 04 2005.

\bibitem{humans-cognitive-map}
R.~Epstein, E.~Z. Patai, J.~Julian, and H.~Spiers, ``The cognitive map in humans: Spatial navigation and beyond,'' {\em Nature Neuroscience}, vol.~20, pp.~1504--1513, 10 2017.

\bibitem{mcts}
C.~B. Browne, E.~Powley, D.~Whitehouse, S.~M. Lucas, P.~I. Cowling, P.~Rohlfshagen, S.~Tavener, D.~Perez, S.~Samothrakis, and S.~Colton, ``A survey of monte carlo tree search methods,'' {\em IEEE Transactions on Computational Intelligence and AI in Games}, vol.~4, no.~1, pp.~1--43, 2012.

\bibitem{World_Model}
D.~Ha and J.~Schmidhuber, ``World models,'' {\em CoRR}, vol.~abs/1803.10122, 2018.

\bibitem{AIF_learning}
K.~Friston, T.~FitzGerald, F.~Rigoli, P.~Schwartenbeck, J.~O. Doherty, and G.~Pezzulo, ``Active inference and learning,'' {\em Neuroscience and Biobehavioral Reviews}, vol.~68, pp.~862--879, 2016.

\bibitem{warehouse}
{aws-robotics}, ``aws-robomaker-small-warehouse-world,'' 2020.
\newblock Accessed: 2024-08-01.

\bibitem{ghost_node_explo}
D.~S. Chaplot, R.~Salakhutdinov, A.~Gupta, and S.~Gupta, ``Neural topological {SLAM} for visual navigation,'' {\em CoRR}, vol.~abs/2005.12256, 2020.

\bibitem{world_model_and_inference}
K.~Friston, R.~J. Moran, Y.~Nagai, T.~Taniguchi, H.~Gomi, and J.~Tenenbaum, ``World model learning and inference,'' {\em Neural Networks}, vol.~144, pp.~573--590, 2021.

\bibitem{ros2}
ROS2, ``Ros2 humble,'' 2022.
\newblock Accessed: 2025-05-21.

\bibitem{turtlebots}
{Turtlebot}, ``Turtlebot versions,'' 2024.
\newblock Accessed: 2024-12-16.

\bibitem{rosbotxl}
Husarion, ``rosbotxl,'' 2025.
\newblock Accessed: 2025-05-21.

\bibitem{nav2}
{nav2}, ``nav2,'' 2021.
\newblock Accessed: 2024-12-01.

\bibitem{potential_field}
Y.~Koren and J.~Borenstein, ``Potential field methods and their inherent limitations for mobile robot navigation,'' vol.~2, pp.~1398 -- 1404 vol.2, 05 1991.

\bibitem{house}
{aws-robotics}, ``aws-robomaker-small-house-world,'' 2021.
\newblock Accessed: 2024-10-01.

\bibitem{nav2_slam}
P.~Gyanani, M.~Agarwal, R.~Osari, {\em et~al.}, ``Autonomous mobile vehicle using ros2 and 2d-lidar and slam navigation,'' {\em Research Square}, vol.~Preprint (Version 1), May 2024.
\newblock Available at Research Square.

\bibitem{FAEL}
J.~Huang, B.~Zhou, Z.~Fan, Y.~Zhu, Y.~Jie, L.~Li, and H.~Cheng, ``Fael: Fast autonomous exploration for large-scale environments with a mobile robot,'' {\em IEEE Robotics and Automation Letters}, vol.~8, pp.~1667--1674, 2023.

\bibitem{ufomap}
D.~Duberg and P.~Jensfelt, ``{UFOMap}: An efficient probabilistic {3D} mapping framework that embraces the unknown,'' {\em IEEE Robotics and Automation Letters}, vol.~5, no.~4, pp.~6411--6418, 2020.

\bibitem{gbplanner}
T.~Dang, M.~Tranzatto, S.~Khattak, F.~Mascarich, K.~Alexis, and M.~Hutter, ``Graph-based subterranean exploration path planning using aerial and legged robots,'' {\em Journal of Field Robotics}, vol.~37, no.~8, pp.~1363--1388, 2020.
\newblock Wiley Online Library.

\bibitem{Darpa_winners}
M.~Tranzatto, M.~Dharmadhikari, L.~Bernreiter, M.~Camurri, S.~Khattak, F.~Mascarich, P.~Pfreundschuh, D.~Wisth, S.~Zimmermann, M.~Kulkarni, V.~Reijgwart, B.~Casseau, T.~Homberger, P.~D. Petris, L.~Ott, W.~Tubby, G.~Waibel, H.~Nguyen, C.~Cadena, R.~Buchanan, L.~Wellhausen, N.~Khedekar, O.~Andersson, L.~Zhang, T.~Miki, T.~Dang, M.~Mattamala, M.~Montenegro, K.~Meyer, X.~Wu, A.~Briod, M.~Mueller, M.~Fallon, R.~Siegwart, M.~Hutter, and K.~Alexis, ``Team cerberus wins the darpa subterranean challenge: Technical overview and lessons learned,'' 2022.

\bibitem{voxblox}
H.~Oleynikova, Z.~Taylor, M.~Fehr, J.~I. Nieto, and R.~Siegwart, ``Voxblox: Building 3d signed distance fields for planning,'' {\em CoRR}, vol.~abs/1611.03631, 2016.

\bibitem{goal_grid_cell}
U.~M. Erdem and M.~Hasselmo, ``A goal-directed spatial navigation model using forward trajectory planning based on grid cells,'' {\em European Journal of Neuroscience}, vol.~35, no.~6, pp.~916--931, 2012.

\bibitem{inductive_AIF}
K.~J. Friston, T.~Salvatori, T.~Isomura, A.~Tschantz, A.~Kiefer, T.~Verbelen, M.~Koudahl, A.~Paul, T.~Parr, A.~Razi, B.~Kagan, C.~L. Buckley, and M.~J.~D. Ramstead, ``Active inference and intentional behaviour,'' 2023.

\bibitem{overcome_explo}
M.~Cai, E.~Aasi, C.~Belta, and C.-I. Vasile, ``Overcoming exploration: Deep reinforcement learning for continuous control in cluttered environments from temporal logic specifications,'' {\em IEEE Robotics and Automation Letters}, vol.~8, no.~4, pp.~2158--2165, 2023.

\bibitem{build_cogn_map}
J.~C.~R. Whittington, D.~McCaffary, J.~J.~W. Bakermans, and T.~E.~J. Behrens, ``How to build a cognitive map,'' {\em Nature Neuroscience}, vol.~25, pp.~1257--1272, October 2022.
\newblock Epub 2022 Sep 26.

\bibitem{AIF_modeling_struct}
R.~Smith, P.~Schwartenbeck, T.~Parr, and K.~J. Friston, ``An active inference approach to modeling structure learning: Concept learning as an example case,'' {\em Frontiers in Computational Neuroscience}, vol.~14, 2020.

\bibitem{bayesian_model_reduc}
K.~Friston, T.~Parr, and P.~Zeidman, ``Bayesian model reduction,'' 2019.

\bibitem{ours_flex_Cmap}
D.~de~Tinguy, T.~Verbelen, and B.~Dhoedt, ``Exploring and learning structure: Active inference approach in navigational agents,'' 2024.

\bibitem{object_nav_semantic}
D.~S. Chaplot, D.~Gandhi, A.~Gupta, and R.~Salakhutdinov, ``Object goal navigation using goal-oriented semantic exploration,'' 2020.

\bibitem{ours_hierarchy}
{de Tinguy, Daria and Van de Maele, Toon and Verbelen, Tim and Dhoedt, Bart}, ``{Spatial and temporal hierarchy for autonomous navigation using active inference in minigrid environment},'' {\em {ENTROPY}}, vol.~{26}, no.~{1}, p.~{32}, {2024}.

\end{thebibliography}
%

\end{document}